\documentclass[sigconf]{acmart}

\usepackage{booktabs} 
\usepackage{enumitem}
\usepackage{listings}
\usepackage{color}

\definecolor{commentgreen}{RGB}{2,112,10}

\newcommand{\cilantro}{\texttt{cilantro}}

\begin{document}

\copyrightyear{2018}
\acmYear{2018}
\setcopyright{acmlicensed}
\acmConference[MM '18]{2018 ACM Multimedia Conference}{October 22--26, 2018}{Seoul, Republic of Korea}
\acmBooktitle{2018 ACM Multimedia Conference (MM '18), October 22--26, 2018, Seoul, Republic of Korea}
\acmPrice{15.00}
\acmDOI{10.1145/3240508.3243655}
\acmISBN{978-1-4503-5665-7/18/10}

\fancyhead{}

\title{\cilantro: A Lean, Versatile, and Efficient Library for Point Cloud Data Processing}

\author{Konstantinos Zampogiannis}
\affiliation{%
  \institution{University of Maryland\\Computer Science Department}
  \streetaddress{4403 A.V. Williams Bldg.}
  \city{College Park}
  \state{Maryland}
  \postcode{20742}
}
\email{kzampog@cs.umd.edu}

\author{Cornelia Ferm\"uller}
\affiliation{%
  \institution{University of Maryland\\Inst. for Advanced Computer Studies}
  \streetaddress{4459 A.V. Williams Bldg.}
  \city{College Park}
  \state{Maryland}
  \postcode{20742}
}
\email{fer@umiacs.umd.edu}

\author{Yiannis Aloimonos}
\affiliation{%
  \institution{University of Maryland\\Computer Science Department}
  \streetaddress{4475 A.V. Williams Bldg.}
  \city{College Park}
  \state{Maryland}
  \postcode{20742}
}
\email{yiannis@cs.umd.edu}

\renewcommand{\shortauthors}{K. Zampogiannis et al.}

\begin{abstract}
We introduce \cilantro, an open-source C++ library for geometric and general-purpose point cloud data processing. The library provides functionality that covers low-level point cloud operations, spatial reasoning, various methods for point cloud segmentation and generic data clustering, flexible algorithms for robust or local geometric alignment, model fitting, as well as powerful visualization tools.
To accommodate all kinds of workflows, \cilantro\ is almost fully templated, and most of its generic algorithms operate in arbitrary data dimension. At the same time, the library is easy to use and highly expressive, promoting a clean and concise coding style. \cilantro\ is highly optimized, has a minimal set of external dependencies, and supports rapid development of performant point cloud processing software in a wide variety of contexts.

\textbf{Availability:} the project source code, with usage examples and sample data, can be found at \url{https://github.com/kzampog/cilantro}.
\end{abstract}

\maketitle

\section{Introduction}\label{sec:intro}
Processing geometric input plays a crucial role in a number of machine perception scenarios. Robots use stereo cameras or depth sensors to create 3D models of their environment and/or the objects with which they interact. Autonomous mobile agents are typically equipped with LiDAR sensors to map their surroundings, localize themselves, and avoid collisions. Consumer electronics are increasingly adopting the integration of depth cameras to identify users and enable ``natural'' user interfaces. The output signals of these sensors are either inherently or directly convertible to 2D or 3D point clouds, highlighting the need for usable and efficient tools for processing raw geometric data.

Thanks to the central role of geometry in the fields of computational theory and computer graphics, a number of notable 3D data processing open-source software libraries have been developed. The Computational Geometry Algorithms Library (CGAL) \cite{cgal:eb-18a} implements algorithms and data structures for an extensive set of tasks, including triangulations, shape analysis, meshing, and various geometric decompositions. The Visualization and Computer Graphics (VCG) library \cite{libvcg} provides a collection of tools for processing and visualizing 3D meshes and constitutes the back-end of the popular MeshLab \cite{cignoni2008meshlab} GUI-based mesh editor. To simplify working with mesh data, libigl \cite{libigl} drops complex data structures in favor of raw data matrices and vectors for shape representations, and supports computation of discrete differential quantities and operators, shape deformation, and remeshing functionalities. While mature and feature-rich in their respective contexts, these software packages have had limited adoption by the machine perception and engineering communities.

The Point Cloud Library (PCL) \cite{Rusu_ICRA2011_PCL} was introduced to fill this gap and became the standard for unorganized point cloud processing among roboticists and machine vision practitioners. It implements numerous algorithms for filtering, feature extraction, geometric registration, reconstruction, segmentation, and model fitting. Partly due to its templated nature, PCL exhibits a steep learning curve; its user-friendliness is further affected by its verbose coding style and typically long application compilation times. Furthermore, its performance in very common tasks is lacking by today's standards (e.g., due to lack of parallelization in many of its modules) and the project has been in an essentially dormant state for a long time.
The Open3D \cite{Zhou2018} library was recently introduced as a potential alternative. It implements 3D point cloud primitive operations (neighbor queries, downsampling, surface normal estimation), as well as useful tools for geometric registration and 3D reconstruction. The library is easy to use and performs better than PCL in common tasks. However, its supported functionality is quite limited, as it essentially only targets 3D reconstruction workflows.

In this paper, we introduce \cilantro, a versatile, easy to use, and efficient C++ library for generic point cloud data processing. We have implemented a concise set of algorithms that cover primitive point cloud operations, spatial reasoning based on convex polytopes, various methods for point cloud segmentation and generic data clustering, flexible algorithms for both robust and local (iterative) geometric alignment, model fitting, as well as powerful visualization tools. The library is written in C++11, is highly templated and optimized, and makes efficient use of multi-threading for computationally demanding tasks. Significant effort has been put into making every component customizable by the user, so that the library does not hinder the development of ``non-standard" workflows. At the same time, we provide useful convenience functions and aliases for the most common algorithm variants, ensuring that our adaptable design does not get in the way of productivity.

By virtue of its flexibility, ease of use, and comprehensive out-of-the-box functionality, \cilantro\ is a great fit for a wide spectrum of workflows, enabling the rapid development of performant point cloud processing software. In the following, we briefly state our main design principles (Section \ref{sec:design}), give a more detailed overview of our supported functionality (Section \ref{sec:functionality}), and demonstrate \cilantro's performance in some ordinary tasks in comparison to equivalent implementations in PCL and Open3D (Section \ref{sec:performance}).

\section{Design overview}\label{sec:design}
We outline \cilantro's main design principles and how they are reflected on the library implementation.

\textbf{Simple data representations.}
We rely on \texttt{Eigen} \cite{eigenweb} matrices and common STL containers to represent point sets and most other entities used/generated by our algorithms. In particular, the input to almost all of our algorithms is a \emph{matrix view} of a point set: a lightweight wrapper, based on the \texttt{Eigen::Map} class template. Our \texttt{ConstDataMatrixMap} universal input template is parameterized by the scalar numerical type (typically \texttt{float} or \texttt{double}) and the point dimensionality (integer value), while dynamic (runtime) dimensionality settings are also supported. The resulting matrix view contains one data point per column. \texttt{ConstDataMatrixMap} is constructible from many common point set representations, such as \texttt{Eigen} matrix variants, STL vectors of fixed-size vector/array objects, raw data pointers, etc. The only requirement is that the mapped data should be contiguously stored in memory, in column-major order. This mechanism is transparent to the user; as a result, users can in most cases directly pass their data to \cilantro\ functions, without the need for type casts or data copying.

\textbf{Generic algorithms.}
Almost all of the algorithms implemented in \cilantro\ operate in arbitrary dimension, according to the nature of the input data. Furthermore, data dimensionality can be either known at compile time or determined and set dynamically.
To accommodate this feature, the library is fully templated, with the exception of visualization and some of the I/O functions.
At the most basic level, template parameters typically include the input data numerical type and dimensionality.
More complex algorithms are adaptable to the user's needs, by being parameterized by the entities responsible for the simpler tasks involved.
For example, our ICP base class only implements an interface for the top-level EM-style iterations of the algorithm and is parameterized by a correspondence search mechanism and an estimator entity.
This way, we can easily generate ICP variants for different metrics (e.g., point-to-point or point-to-plane) and different correspondence search engines (e.g., $k$d-tree based or projective), while maintaining a common, general interface.
For performance reasons, we have opted to implement our ``modular" high-level algorithms as \emph{static} interfaces, by making use of the Curiously Recurring Template Pattern (CRTP) idiom.

\textbf{Ease of use.}
Generality should not come at the cost of usability. For this reason, we tackle template parameter verbosity by providing type aliases and convenience functions for the most standard variants of our algorithms.
We also increase \cilantro's expressiveness by enabling method chaining in almost all of our classes.
Furthermore, the library code is maintained in a clean, consistent style, with class and function names that are descriptive of the underlying functionality.
The following code snippet showcases the library usage in a very simple pipeline that involves reading a 3D point cloud from a file, downsampling it by means of a voxel grid, estimating its surface normals, and saving the result to a file:
\begin{lstlisting}
#include <cilantro/point_cloud.hpp>

int main(int argc, char ** argv) {
    cilantro::PointCloud3f(argv[1]).gridDownsample(0.005f).estimateNormalsRadius(0.02f).toPLYFile(argv[2]);
    return 0;
}
\end{lstlisting}
The code is very concise, with minimal boilerplate. Readers familiar with PCL will reckon that an equivalent PCL implementation would require a significantly larger amount of code.

\textbf{Performance.}
The library is highly optimized and makes use of OpenMP parallelization for computationally demanding operations. 
Significant effort has been put into benchmarking alternative implementations of common operations in order to find the fastest variant and/or parallelization pattern.
As a result, \cilantro\ exhibits the lowest running times in the benchmarks of Section \ref{sec:performance}.

\textbf{Minimal dependencies.}
\cilantro\ was built upon a carefully selected set of third party libraries and has minimal external dependencies. The library comes bundled with: \texttt{nanoflann} \cite{blanco2014nanoflann} for fast $k$d-tree queries, Spectra \cite{spectra}, an ARPACK-inspired library for large scale eigendecompositions, Qhull \cite{barber1996quickhull} for convex hull and halfspace intersection computations, and \texttt{tinyply} \cite{tinyply} for PLY format geometry I/O. The only external dependencies are:
\begin{itemize}[noitemsep,topsep=0pt,leftmargin=*]
\item \texttt{Eigen} \cite{eigenweb}, an elegant and efficient linear algebra library on which we rely for most of our numerical operations, and
\item Pangolin \cite{pangolin}, a lightweight OpenGL viewport manager and image/video I/O abstraction library, on which our visualization modules were built.
\end{itemize}
\cilantro\ uses the CMake build system and can be compiled with all major C++ toolchains (GCC, Clang).

\section{Functionality}\label{sec:functionality}
For most of our supported functionality, we have adopted an object-oriented approach, implementing each of our algorithms as a class template, while also providing free functions for simpler operations.
The library components can be conceptually divided in the following categories.

\textbf{Core operations.}
We support a wide set of primitive point cloud operations by implementing:
\begin{itemize}[noitemsep,topsep=0pt,leftmargin=*]
\item An optimized, user-friendly \texttt{KDTree} template that supports general dimension $k$-NN, radius, and hybrid queries, under all of the distance metrics supported by \texttt{nanoflann}.
\item Arbitrary dimension surface normal and curvature estimation.
\item Principal Component Analysis (PCA).
\item A generic, optimized \texttt{GridAccumulator} template and its appropriate instantiations for general dimension grid-based point cloud downsampling.
\item I/O functions for general matrices and 3D point clouds.
\item Utility functions for RGBD to/from 3D point cloud conversions.
\end{itemize}
All the above algorithms operate on ``raw", \texttt{ConstDataMatrixMap}-wrapped data. To facilitate common workflows, we also provide a convenience \texttt{PointCloud} class template that encapsulates point coordinates, normals, and colors, and provides basic point selection functionality, as well as a large number of helper methods for normal estimation, downsampling, and geometric transformations. In the 3D case, \texttt{PointCloud} instances are directly constructible from PLY files and RGBD image pairs.

\textbf{Spatial reasoning.}
Building on Qhull's facilities and a simple feasibility solver, we provide a \texttt{ConvexPolytope} template that is constructed as either the convex hull of an input point set or the halfspace intersection defined by an input set of linear inequality constraints, enabling seamless representation switching.
Our \texttt{SpaceRegion} class represents arbitrary space regions as unions of convex polytopes. 
Both space representations implement set operations (intersection for \texttt{ConvexPolytope}, all common ones for \texttt{SpaceRegion}), interior point tests, volume calculation, and can be transformed geometrically.

\textbf{Clustering/segmentation.}
\cilantro\ provides four standard clustering algorithms: a parallel, optimized $k$-means implementation that can optionally use $k$d-trees for large numbers of clusters, three common spectral clustering variants \cite{von2007tutorial} that rely on Spectra, an arbitrary kernel mean-shift implementation, and a parallelized, generic connected component segmentation algorithm that supports arbitrary point-wise similarity functions. We note that common segmentation tasks such as extracting Euclidean (see PCL's \texttt{EuclideanClusterExtraction}) or smooth (see PCL's \texttt{RegionGrowing}) segments can be straightforwardly cast as connected component segmentation under different similarity metrics.

\textbf{Iterative geometric registration.}
We provide a CRTP base class template that implements the top-level loop logic (alternating between correspondence estimation and transformation parameter optimization) of the Iterative Closest Point (ICP) algorithm. The base template is parameterized by the correspondence estimation mechanism and the transform estimator. We implement a standard, $k$d-tree based correspondence search engine, which can operate on arbitrary point feature spaces, as well as one for correspondences by projective data association for the 3D case. We provide optimizers for both \emph{rigid} \cite{besl1992method,low2004linear} and \emph{non-rigid} \cite{sorkine2007rigid,sumner2007embedded} (by means of a robustly regularized, locally rigid warp field) pairwise registration under the the point-to-point and point-to-plane metrics, as well as weighted combinations thereof. Our modular design can be easily extended to accommodate less common ICP processes (e.g., for multi-way registration). At the same time, \cilantro\ provides useful shortcut wrappers/definitions for the more common registration workflows.

\textbf{Robust estimation.}
The library comes with a RANSAC template that is meant to be used as a CRTP base class and only implements the top-level (random sampling/inlier evaluation) loop logic. We currently provide two general dimension RANSAC estimator instances: one for robust (hyper)plane fitting and one for rigid pairwise point cloud alignment given (noisy) point correspondences.

\textbf{Visualization.}
\cilantro\ implements a fully customizable, interactive, and easy to use 3D \texttt{Visualizer} class that supports: an extensible set of renderable entities with adjustable rendering options, complete control over all projection parameters, both perspective and orthographic projection modes, image capturing and video recording of the viewport, and custom keyboard callbacks.
We also provide a convenience \texttt{ImageViewer} class.
Both our visualization facilities are based on Pangolin.
To facilitate the visualization of data of potentially unknown dimension, for which we are only given pairwise distances, we have included a classical Multidimensional Scaling (MDS) implementation that can be used to compute low dimensional, distance-preserving Euclidean embeddings.

\begin{figure}[!ht]
  \centering
  \includegraphics[width=0.495\columnwidth]{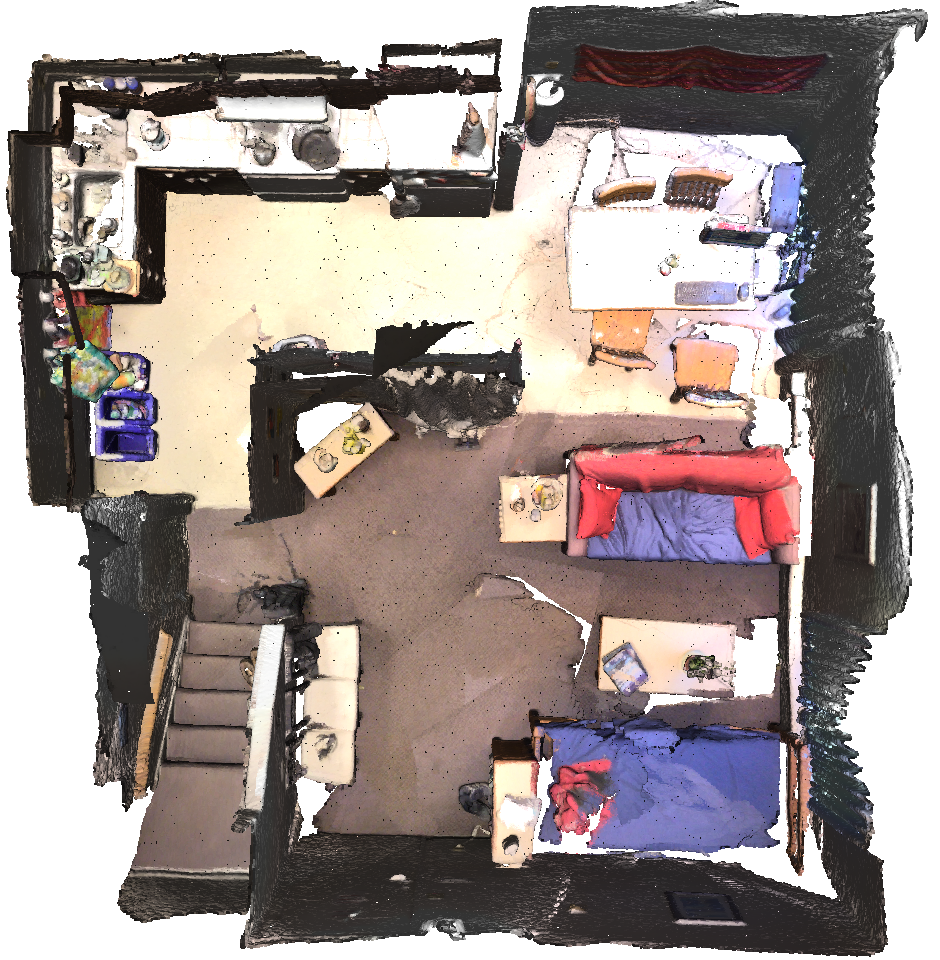}
    \includegraphics[width=0.495\columnwidth]{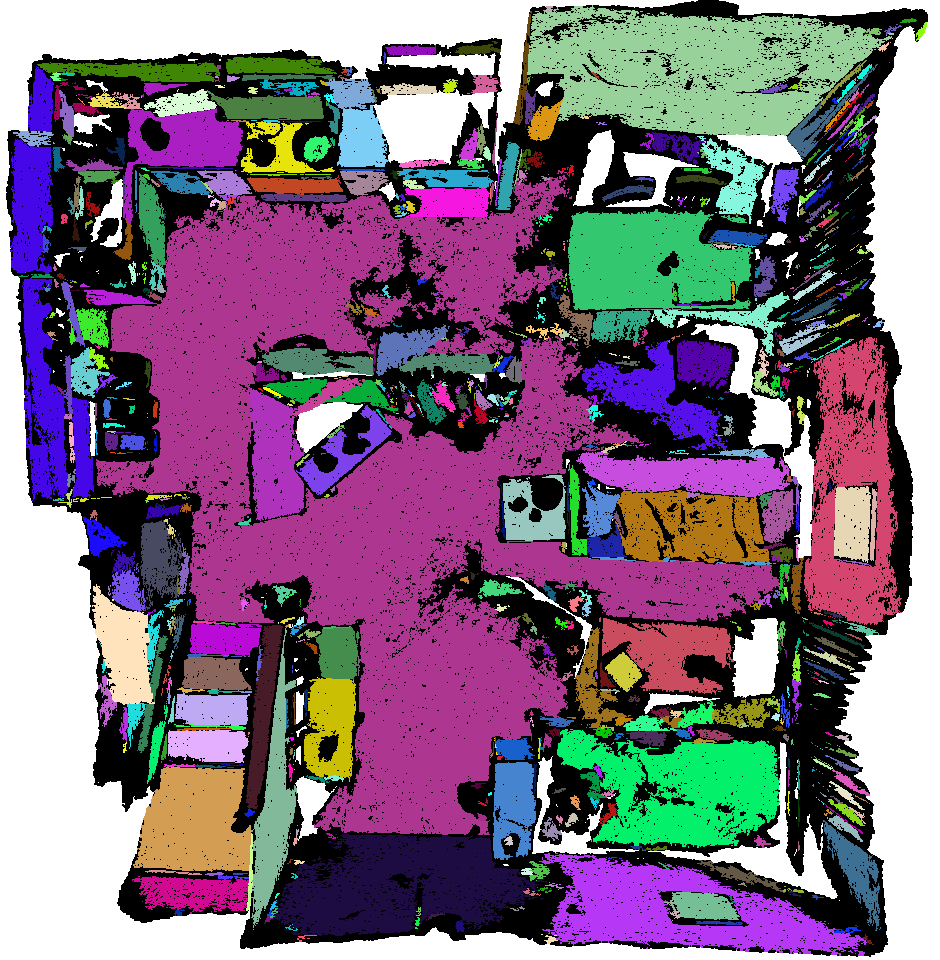}
    \vspace{-2em}
    \caption{Point cloud segmentation into smooth segments. Black points correspond to very small clusters that were discarded.}
    \label{fig:ccs}
\end{figure}

\section{Performance}\label{sec:performance}
We compare \cilantro's performance in common 3D point cloud processing tasks against PCL and Open3D.
Our benchmarking was done on a standard Linux (Ubuntu 16.04 based) desktop machine, equipped with an Intel\textregistered\ Core i7-6700K CPU (4 cores, 8 threads).
All three libraries were compiled from source in `Release' configuration, with compiler (GCC 5) optimizations set to the highest level (\texttt{-O3}).

Our first test involves the segmentation of an apartment-scale 3D reconstruction into smooth segments using PCL's \texttt{RegionGrowing} and \cilantro's \texttt{ConnectedComponentSegmentation}.
We use the \texttt{apt0} model (Fig. \ref{fig:ccs}, left) as input, available on the BundleFusion \cite{dai2017bundlefusion} project website, which consists of roughly 7.8 million points.
After downsampling the input using a voxel grid of bin size equal to 5mm, using $k$NN neighborhoods with $k=30$ and a normal angle threshold of $2.8$ radians, \cilantro\ produces the result of Fig. \ref{fig:ccs} (right) in 3.65 seconds, while, for the same parameters, the PCL implementation took 17.34 seconds. Open3D offers no equivalent/similar functionality.

PCL's standard point types use single precision floating point numbers, while Open3D works only with double precision. For all subsequent computations, we use double precision for \cilantro.

\begin{figure}[!ht]
    \centering
    \vspace{-1em}

    \includegraphics[width=1.00\columnwidth]{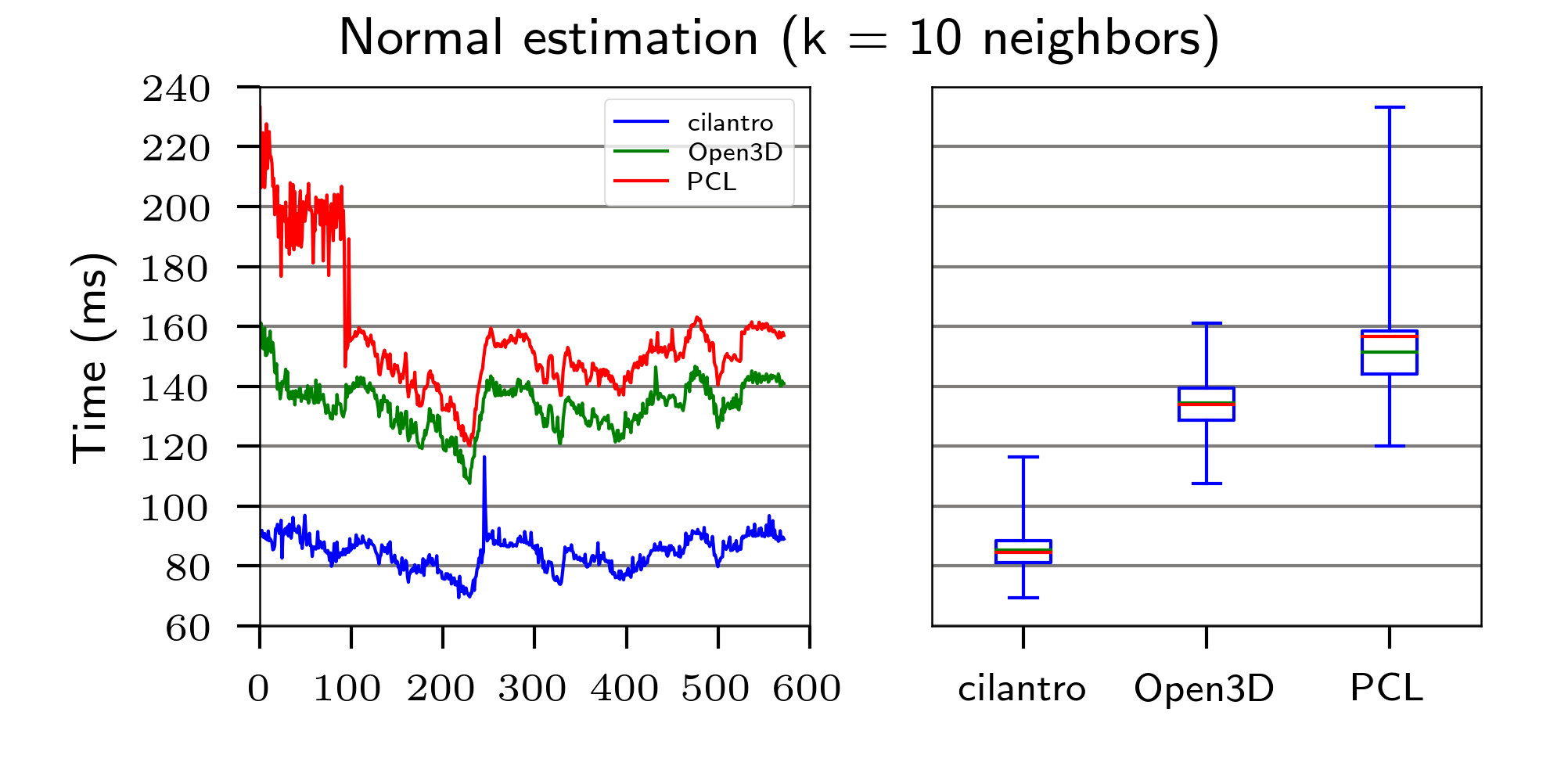}
    \vspace{-1.875em}

    \includegraphics[width=1.00\columnwidth]{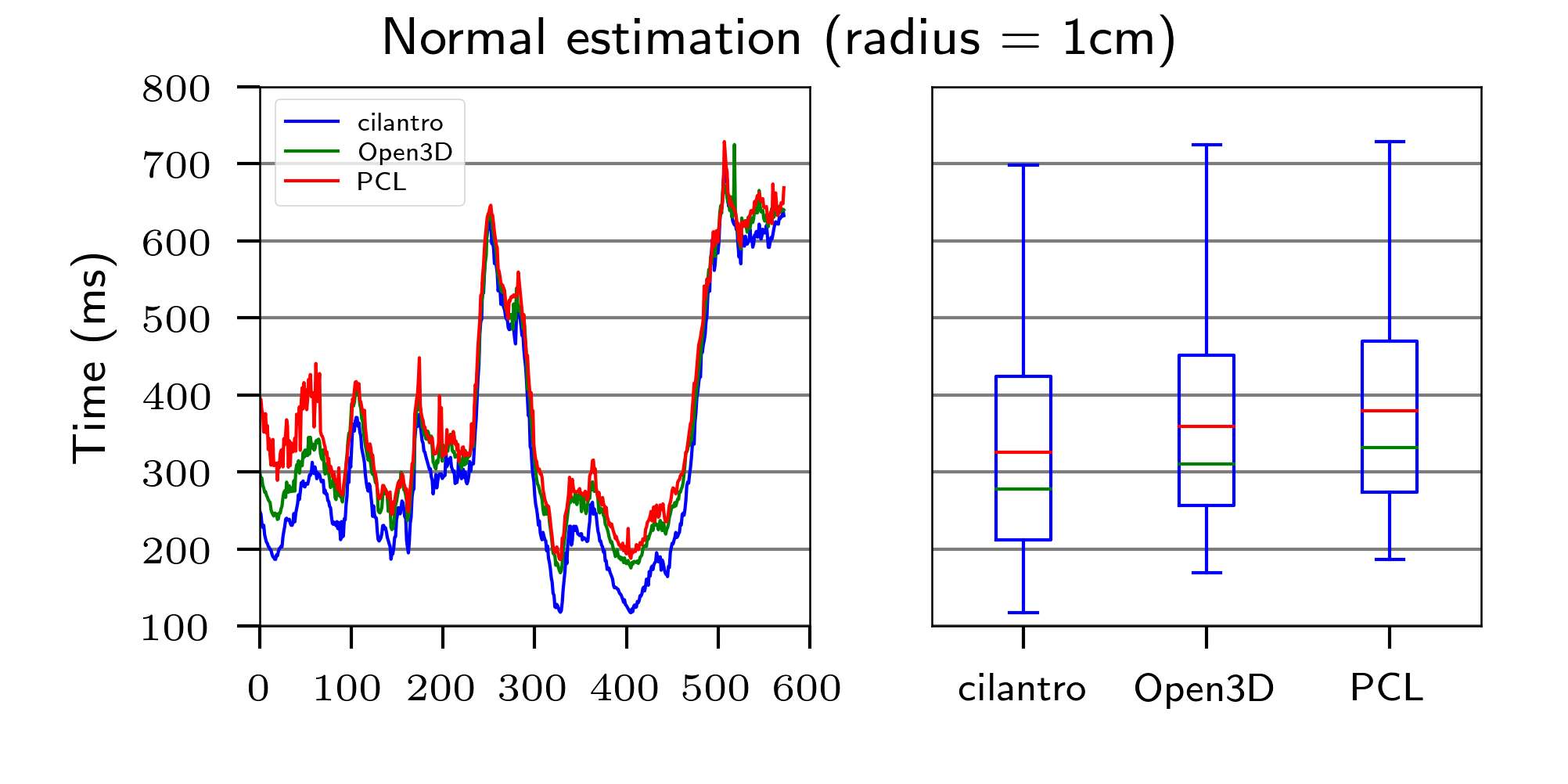}
    \vspace{-1.875em}

    \includegraphics[width=1.00\columnwidth]{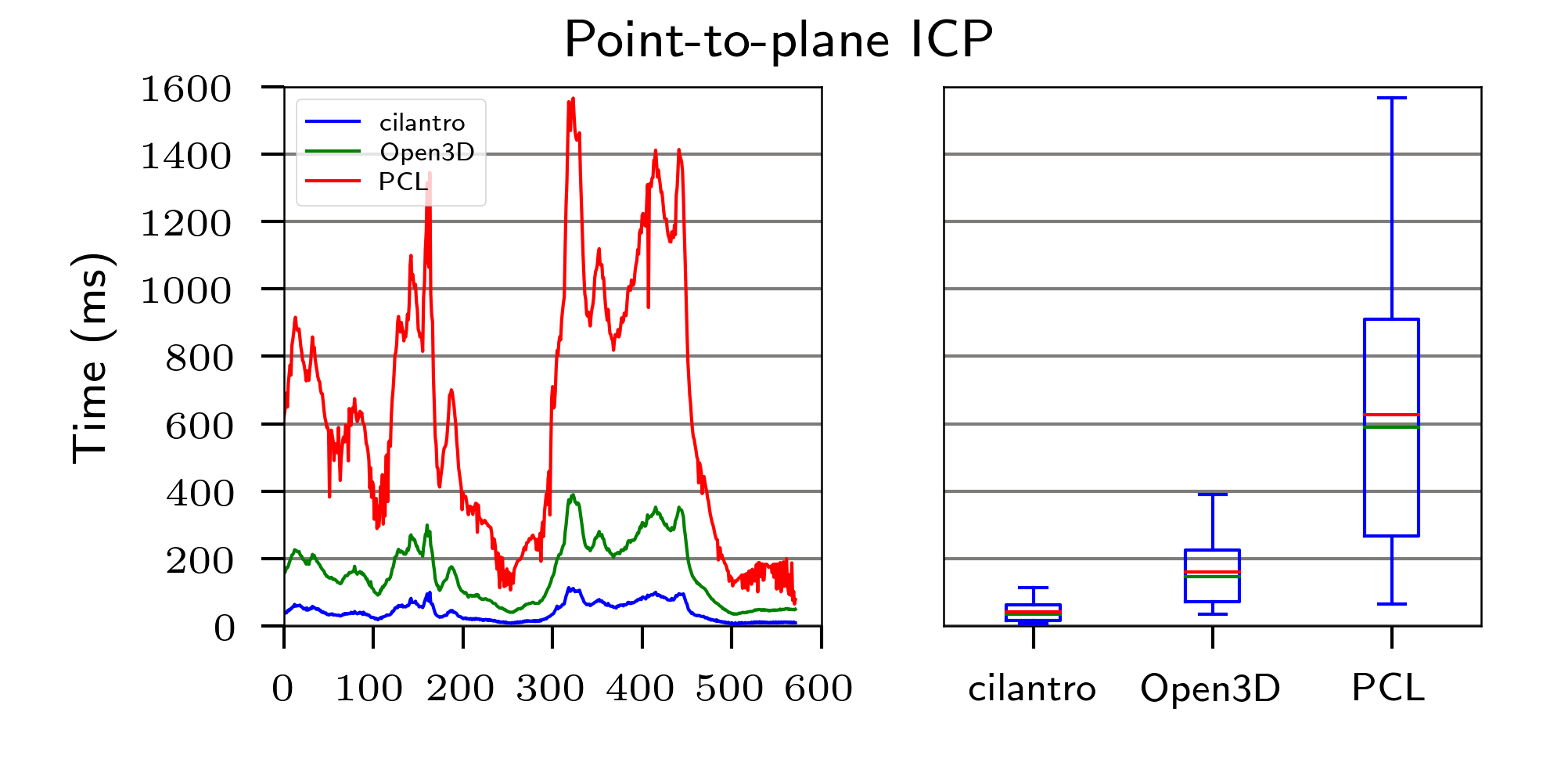}
    \vspace{-3em}

    \caption{Performance comparisons against PCL and Open3D in common operations. Left column: running time as a function of the input video frame. Right column: box and whisker plots of running times per library (red lines are means, green lines are medians).}
    \label{fig:performance}
\end{figure}

In our next test, we compare our normal estimation performance against equivalent implementations in Open3D (parallelized by default) and PCL (the \texttt{NormalEstimationOMP} variant).
We use the \texttt{fr1/desk} sequence of the TUM RGBD dataset \cite{sturm12iros} as input (about 600 pre-registered RGBD image pairs at VGA resolution).
We run all three implementations on all input frames at full resolution, using two types of local neighborhoods: one defined by the 10 nearest neighbors and one defined by a radius of 1cm.
As can be seen in the two top rows of Fig. \ref{fig:performance}, \cilantro\ consistently outperforms the other two implementations. In the $k$NN case in particular, it exhibits speedups of $1.58\times$ and $1.85\times$ over Open3D and PCL, respectively.
We note that \cilantro\ and PCL enforce viewpoint-consistent normals during estimation, while in Open3D this has to be done in a separate pass, which was not performed in our experiments.

In our last test, we compare point-to-plane ICP performance among the three libraries. We used the same \texttt{fr1/desk} sequence as before, aligning each frame to its previous one. All point clouds were downsampled using a voxel grid of bin size equal to 1cm, and normals were estimated based on 10 nearest neighbor local neighborhoods. To account for different termination criteria conventions, we forced all three implementations to perform a fixed number of 15 iterations. In all three cases, correspondences were established by means of nearest neighbor $k$d-tree queries, with a rejection distance threshold of 5cm.
Registration times are reported in the third row of Fig. \ref{fig:performance}.
We note that, at all times, all three implementations converged to the same transformation, up to numerical precision.
\cilantro's ICP implementation outperforms the other two by a significant margin, demonstrating speedups of $3.82\times$ and $14.99\times$ over Open3D and PCL, respectively.

\section{Software release}\label{sec:release}
\cilantro\ is released under the MIT license and its source code is openly available at \url{https://github.com/kzampog/cilantro}. Contributions from the open-source community are welcome, via the GitHub issues/pull request mechanisms.

The library is under active, continuous development, undergoing frequent API improvements, functionality additions, and performance optimizations. Future additions will include structures and algorithms for surface merging/reconstruction, more point cloud resampling strategies, and support for geometry I/O in other file formats.
We are also investigating GPU implementations for some of our algorithms, focusing on improving our non-rigid registration performance.

\begin{acks}
The authors would like to thank Aleksandrs Ecins and Kanishka Ganguly for their code contributions and support, and Moschoula Pternea for generating the plots in this report.
The support of ONR under grant award N00014-17-1-2622 and the support of the National Science Foundation under grants SMA 1540916 and CNS 1544787 are greatly acknowledged.
\end{acks}

\bibliographystyle{ACM-Reference-Format}
\bibliography{references}

\end{document}